\documentclass[sigplan,screen,nonacm,peerreviewca,pbalance]{acmart}

\usepackage{caption}
\usepackage{amsmath,amsfonts}
\usepackage{algorithmic}
\usepackage{textcomp}
\usepackage{xcolor}
\usepackage{soul}   
\usepackage{xparse}
\usepackage{tabularx}
\usepackage{subcaption}     
\usepackage{listings}   
\usepackage{booktabs}  
\usepackage{array}     
\usepackage[export]{adjustbox}

\usepackage{placeins}   

\def\BibTeX{{\rm B\kern-.05em{\sc i\kern-.025em b}\kern-.08em
    T\kern-.1667em\lower.7ex\hbox{E}\kern-.125emX}}
\usepackage{graphicx}
\usepackage[switch]{lineno}

\lstdefinestyle{Cstyle}{
  backgroundcolor=\color{gray!10},   
  basicstyle=\ttfamily\scriptsize,        
  keywordstyle=\color{blue},         
  stringstyle=\color{red!60!brown},  
  commentstyle=\color{green!50!black},
  frame=single,                      
  tabsize=2,
  showstringspaces=false,
  breaklines=true,
  xleftmargin=0em,          
  framexleftmargin=0.1em     
}
\newcommand{\code}[1]{\texttt{\small #1}}

\title{T-LLM Compiler: Trusted LLM-based Code Optimization and Verification Framework}

\author{Zahra Fazel}
\email{zahra.fazel@huawei.com}
\affiliation{%
  \institution{Huawei Technologies, Heterogeneous Compiler Lab}
  \city{Toronto}
  \country{Canada}
  }
  
\author{Sunanda Gamage}
\email{sunanda.gamage@huawei.com}
\affiliation{%
  \institution{Huawei Technologies, Heterogeneous Compiler Lab}
  \city{Toronto}
  \country{Canada}
  }
  
\author{Shayan Shirahmad Gale Bagi}
\email{shayan.shirahmad.bagi@huawei.com}
\orcid{0000-0001-5773-3799}
\affiliation{%
  \institution{Huawei Technologies, Heterogeneous Compiler Lab}
  \city{Toronto}
  \country{Canada}
  }
  
\author{Amir H. Ashouri}
\email{amirh.ashouri@gmail.com}
\orcid{0000-0001-8606-6497}
\affiliation{%
  \institution{Huawei Technologies, Heterogeneous Compiler Lab}
  \city{Toronto}
  \country{Canada}
  }

\author{Tomasz S. Czajkowski}
\email{tomasz.czajkowski@huawei.com}
\orcid{0009-0008-3294-6198}
\affiliation{%
  \institution{Huawei Technologies, Heterogeneous Compiler Lab}
  \city{Toronto}
  \country{Canada}
  }  
  
\author{Bryan Chan}
\email{bryan.chan@huawei.com}
\orcid{orcid.org/0009-0004-7389-5017}
\affiliation{%
  \institution{Huawei Technologies, Heterogeneous Compiler Lab}
  \city{Toronto}
  \country{Canada}
  }
  
\author{Reza Azimi}
\email{reza.azimi1@huawei.com}
\orcid{0009-0006-6000-8997}
\affiliation{%
  \institution{Huawei Technologies, Heterogeneous Compiler Lab}
  \city{Toronto}
  \country{Canada}
  }
    
\author{Yaoqing Gao}
\email{yaoqing.gao@huawei.com}
\orcid{0000-0002-5392-5088}
\affiliation{%
  \institution{Huawei Technologies, Heterogeneous Compiler Lab}
  \city{Toronto}
  \country{Canada}
  }

\begin{document}


\begin{abstract}
Recent advances in Large Language Models (LLMs) have opened opportunities to apply high-level code transformations to the field of code optimization, and it has since emerged as one of the most fundamental tasks for LLMs to perform; however, at present, LLMs struggle to apply wide-ranging code optimization tasks due to both the complexity of the code and the inability to independently verify the correctness of the transformations. In this paper, we present the Trusted LLM (T-LLM) Compiler, which proposes an advancement in compiler technology through a collaborative effort involving high-level LLM code transformations, traditional compilers, and verification tools. Experimental results reveal that it can significantly improve code correctness when tested on a set of PolyBench/C benchmarks. Our approach facilitates iterative code optimization efforts with verification strategies that enable corrective actions. Through this approach, T-LLM Compiler achieves code optimization accuracy of up to 83.3\% and a speedup of up to 16.1\% on the PolyBench/C benchmarks, with the transformed code reaching an average of 26.7\% speedup wrt standard baselines. Additionally, we release the project's source code to the open-source community \footnote{\url{https://github.com/BiSheng-Compiler-Agents/CCE-WOZ/tree/tllm-compiler}}. 

\end{abstract}

\maketitle


\section{Introduction}
During compilation, program optimization is an important task for improving application performance and usability. The task requires substantial knowledge of algorithms, computer architecture, and the underlying platform on which the application runs. However, because of the high degree of expertise and low-level knowledge required, it becomes a challenging task to automate. Since the early days of programming, compilers have emerged as core tools to first transform code into executable instructions, and later, with the advent of optimizing compilers, to actually perform mid-to-low-level transformations that reduce the runtime, power, and code size of any given application. Compilers, however, are limited in what they can do in part because of programming languages that impose certain restrictions on code written, as well as the increasing complexity of low-level code transformations required to make significant improvements.

To address this problem, numerous approaches have been proposed that leverage artificial intelligence (AI), be it through the use of machine learning (ML) models \cite{trofin2021mlgo,ashouri2022mlgoperf,ashouri2023acpo,ashouri2026protean}, AI-powered tools such as RL-based autotuners \cite{wang2022automating}, or large language models \cite{cummins2025llm,taneja2025llm}. ML models are generally used to replace a hand-coded heuristic that determines when a certain transformation should be applied or is profitable. This is a very useful approach, as it allows a more data-driven approach to optimization that can easily adapt to new architectures with only model retraining being required, while retaining the correctness of the transformation since that portion is performed by traditional compiler approaches. AI-powered tuners serve a similar purpose, but leverage compiler flags to redirect a search of viable solutions in a particular direction. While the search typically involves sweeping many hundreds of parameters, RL-based approaches have proven to be highly effective in guiding the search. Finally, with the advent of LLMs, a wide range of efforts has emerged to replace some compiler optimizations with LLM source code changes, which can be a viable path to performance improvement. Also, work has been done on low-level virtual machine intermediate representation (LLVM-IR) to apply large language models to predict pass ordering for the best code size optimization sequence or direct IR-to-IR transformations \cite{cummins2025llm}. These approaches, however, have not improved the transformation success rate.

\begin{figure*}[t!]
    \centering
    \includegraphics[trim={0in, 2.1in, 0in, 2in}, clip, width=\textwidth, page=1]{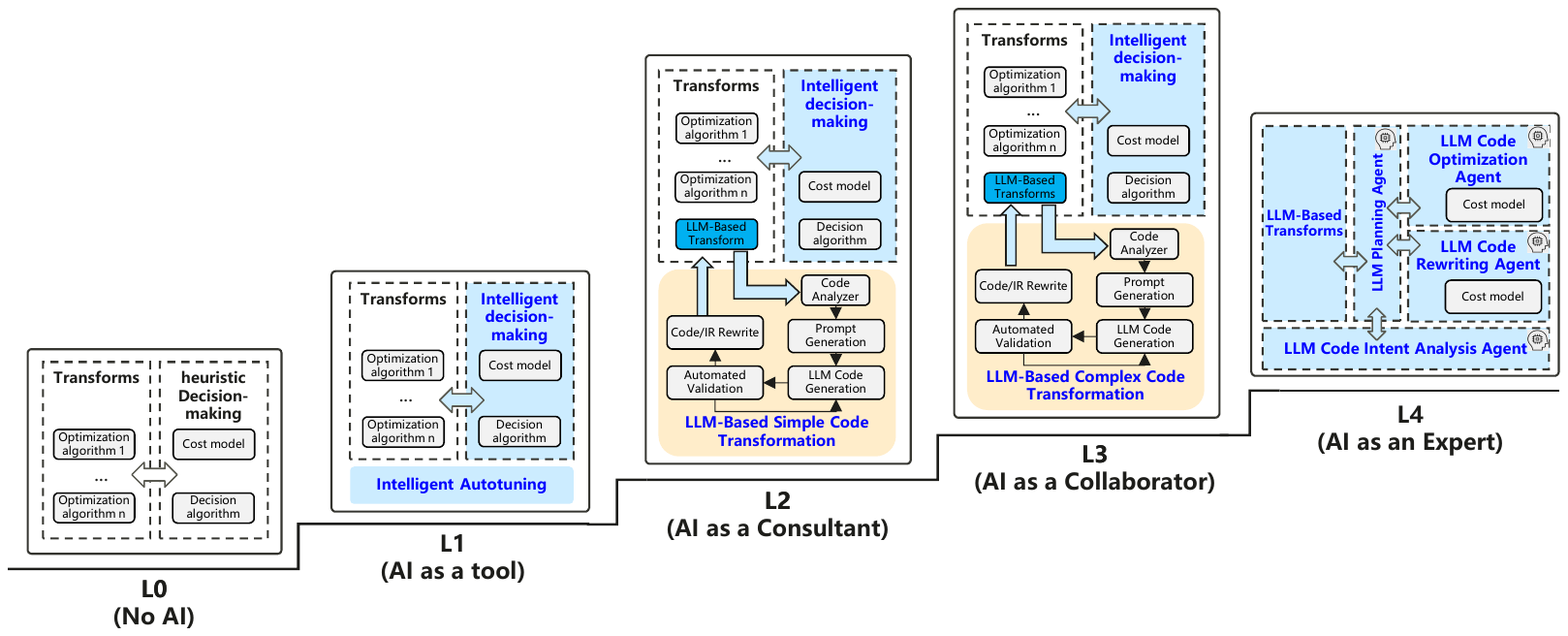}
    \caption{T-LLM Compiler AI-Assistance levels}
    \label{fig:t_llm_compiler_levels}
\end{figure*}

In this paper, we propose that AI can catalyze the advancement of optimizing compilers in a collaborative fashion, where LLMs can unlock performance opportunities in tandem with existing compiler tools. To achieve this, we propose that an increasing level of compiler-AI integration is the key to this next generation of compilers. To illustrate the idea, we have divided compiler AI-enablement into levels inspired by \cite{morris2024position} as shown in Fig. \ref{fig:t_llm_compiler_levels}.

In the figure, level 0 represents a base compiler without any AI, or more specifically, machine learning-based enablement, where the decisions are made by heuristic algorithms. Changes to the compiler require code rewriting and can be time-consuming. Level 1 is basic machine-learning-based enablement, where decisions of profitability can be made by machine learning models that are trained on a large corpus of data, allowing for fine-grained cost function tuning and data-driven compiler design. The transformations themselves are still performed using traditional methods. At level 2, we inject the use of LLMs to rewrite code in a more substantial way, for example, by permitting algorithmic changes that general compilers are not able to do. This, by definition, requires correctness checking, but it provides an opportunity to open up optimization space for compilers. At this stage, we consider AI/LLMs as consultants to help in unlocking the performance or effectiveness of existing compiler passes. This collaborative effort mirrors minor code changes developers would do to boost performance, with some knowledge of where and why optimizing compilers fail to apply the most effective transformations. This approach can be further advanced to level 3, where more complex transformation sequences are entrusted to LLMs but do not replace the compiler fundamentals that are used for otherwise effective optimizations. In this flow, the level 3 compiler integrates LLMs in synergy with traditional optimizing compilers. Finally, level 4 elevates the optimization process to one where a wide range of tools are integrated into the compilation process, allowing for both local and global optimizations to be performed by a combination of LLM and traditional transforms, aided by code analysis tooling to direct optimization effort.

In this paper, we propose the Trusted LLM Compiler (T-LLM Compiler), which approaches level 3 by leveraging various code transformations to unlock compiler capabilities and produce fast code. In this approach, we leverage LLMs to apply optimization strategies and then validate the results using a wide range of verification techniques. This is done in an iterative fashion to generate a correctly optimized program. To ensure correctness, we include a comprehensive validation framework comprising a compiler for syntax checking, as well as LLMs and tools such as Alive2 \cite{lopes2021alive2} and CBMC \cite{kroening2014cbmc}, to address algorithmic correctness challenges. This automated iterative framework leverages the Qwen2.5-32B-Instruct LLM model for code transformations, and together with a validation strategy, shows both high speedup and correctness reaching 83\%, significantly advancing the state-of-the-art in LLM-enabled compiler technology.

In the following sections, we describe key contributions of our work: 1) the T-LLM Compiler framework, detailed design information of the flow and prompting strategies, and 2) our transformation verification framework that comprises syntax, symbolic, and semantic checking approaches to improve the correctness of optimized code.


\section{Related Work}

With the popularity and potential impact LLMs promise, many attempts to improve code optimization have been proposed. Generally, we can categorize them into two groups: deep-learning methods for code optimization and LLM-based code optimization systems.

\subsection{Deep-learning methods for code optimization}

The first category of works focuses on the training and/or fine-tuning of large language models towards a specific task. The idea is to enclose within the LLM the capability to effectively perform a task in a single step on the basis of prior experience gained via training or finetuning. The main thrust in the context of compilation and code optimization tasks is therefore to replace compilers with LLMs, although in many cases, the works still use a compiler as a final step for executable generation.

In the domain of code optimization, a great deal of focus is given to training datasets and methodologies for models. For example, \cite{shypula2023learning} \cite{shypula2025automated} uses C++ programs from CodeNet \cite{puri2021project}, using gem5 \cite{binkert2011gem5} as a simulator to evaluate the performance of each solution to a problem. The data is curated and categorized by problems and used for training and validation purposes. Their proposed dataset, performance-improving edits (PIE), is used to create slow-fast code pairs to help LLMs learn optimization patterns to replicate.

In addition to training, prompt selection has emerged as an important consideration for LLM-based code optimization systems.  Works such as \cite{garg2025rapgen} discuss the challenge of addressing performance bugs and use retrieval-augmented prompt generation to identify similar performance-related bugs that can then be used to retrieve an appropriate prompt that has been proven to successfully guide an LLM in problem resolution, showing that such an approach can be more successful than purely manual approaches 60\% of the time.

In the compiler domain, the LLM Compiler work \cite{cummins2025llm} focuses on training foundational models that take as input LLVM-IR \cite{lattner2004llvm} to drive the generation of optimized code (optimized for code size) and a pass sequence that an LLVM compiler would use to generate the desired outcome. In the case of direct code generation, they achieve successful compilation in 95.6\% of test cases; however, only 20\% matched compiler output (desired), and correctness via runtime checking was not performed. In the scenario of pass inference, the flow used -Oz pass in cases where the generated pass sequence was invalid, and the actual transformation was performed by the LLVM compiler, resulting in a modest 5.58\% code size reduction.

In contrast to optimizing code at the LLVM-IR level, DeepPERF \cite{garg2022deepdev} proposes an approach to suggest C\# code changes for developers to apply. The model is trained on both the English language and source code corpora, followed by fine-tuning for the task of generating patches aimed at performance optimization. 
Similarly, \cite{chen2022learning} proposes a model trained on code pairs to guide programmers in writing high-performance code. 

\subsection{LLM-based code optimization systems}

The limitation of LLM-only approaches is that if the LLM fails to generate the exact result required, the process fails. To address this, other works have developed comprehensive systems around LLMs to enhance the chances of success. These systems usually comprise a form of checking to determine if the output is valid or desirable, which causes a retry or a refinement step in the code optimization process.

LLM-Vectorizer \cite{taneja2025llm} is a recent LLM-based code optimization flow geared toward producing more effective AVX2 vectorized code, tested on the TSVC-2 benchmark set. The framework shows the ability to generate correct vectorized code in ~52\% of the time; however, the success rate increases with an increased number of attempts. This framework utilizes GPT-4 \cite{achiam2023gpt} for code transformations and Alive2 \cite{lopes2021alive2} for verifying the transformed code. Alive2 essentially performs analysis on two input LLVM-IR representations of presumably equivalent code to verify if the provided inputs are indeed equivalent, which is very useful in validating LLVM compiler transforms. However, it is limited in its ability to handle complex loop structures.

ECO \cite{lin2025eco} is another framework that examines code improvements based on performance anti-patterns, or code patterns that indicate a performance opportunity. In this case, the work focuses on issues such as memory allocation and the use of data structures, including vectors and maps. The objective is to propose patches to address these performance bottlenecks, and the results are validated via unit tests and integration tests. LLMs are used to correct some simple failures, but eventually, failing test cases are abandoned. Passing cases are still reviewed by engineers. Similar works include \cite{bairi2024codeplan}\cite{meng2011systematic}\cite{miltner2019fly}.


\section{System Design and Implementation}

\begin{figure*}[ht]
    \centering
    \includegraphics[trim={0in, 2.4in, 0in, 2.3in}, clip, width=\textwidth, page=1]{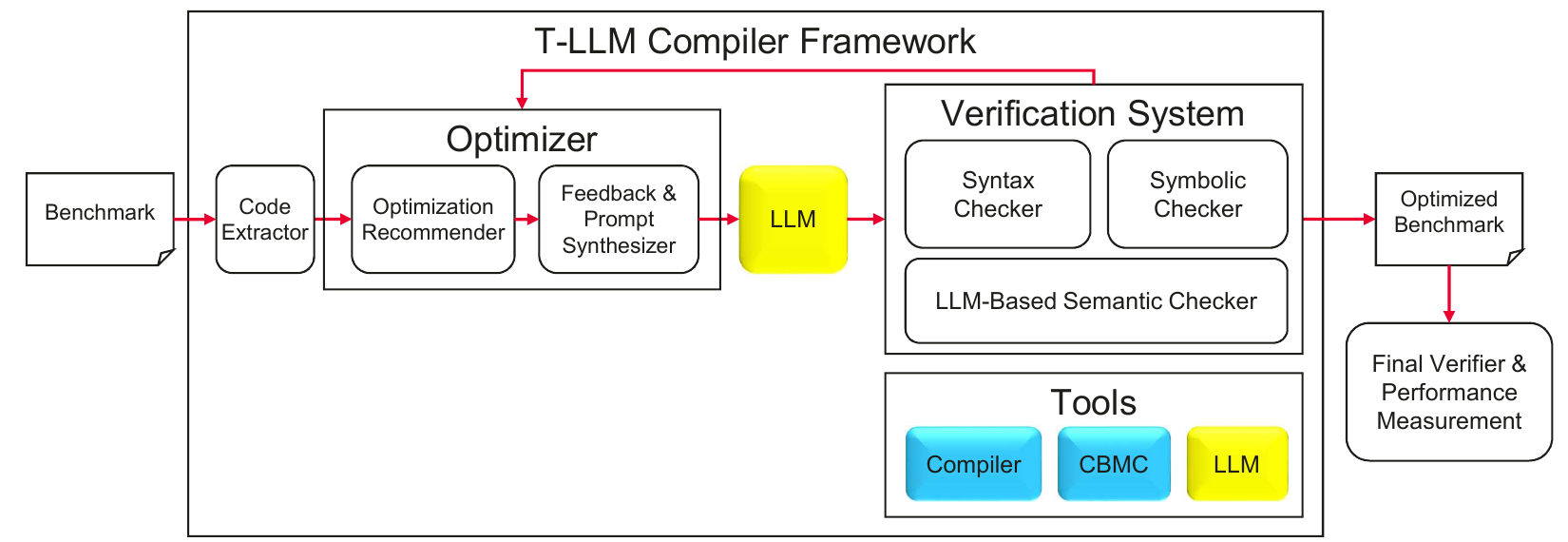}
    \caption{Architecture of our code optimization and verification system. LLMs are shown in yellow, and other tools are shown in blue.}
    \label{fig:system_architecture}
\end{figure*}

To implement our T-LLM compiler, we constructed a system depicted in Fig. \ref{fig:system_architecture}. The system takes as input a benchmark source code and feeds it to the optimizer. Within the optimizer, the first step is to determine if this is a retry or a first attempt at optimization. In the case of a first attempt, which combines constructions and optionally includes examples of transformations, along with the code to be optimized, it passes the instructions to the LLM for optimization. Then, the LLM applies the prompt, which comprises instructions and code to be optimized, and produces the resulting optimized code. The code is then checked using syntax, symbolic and semantic checkers that validate its correctness against the input source code. In the case of failure, the feedback loop to the optimizer is engaged, and corrective actions are taken to fix the code previously generated. This could range from simple syntax fixes to logical corrections detected during  verification. Finally, once a verified code has been generated or the maximum iteration count is reached, the flow exits, producing the proposed optimized program. The resulting output is then tested against the provided test cases, and performance is measured.

\subsection{Optimizer}
The optimizer employs a Qwen2.5-32B-instruct \cite{team2024qwen2} large language model (LLM) to perform code transformations for loop optimization. This process is guided by prompts that define the optimization strategy, leveraging few-shot learning and chain-of-thought techniques to improve reliability and quality of the transformations. The optimizer covers both the original code optimization attempt and the feedback from the verifier. The prompt applied to the LLM comprises the following:
\begin{itemize}
\item Original code to be optimized, together with instructions (based on the recommended prompt) to perform code optimization.
\item Depending on the strategy, the optimizer incorporates few-shot learning or chain-of-thought exemplars, providing the LLM with representative transformations that guide its reasoning process.
\item Feedback from the verification chain to address enumerated errors in the case of verifier rejections.
\end{itemize}

To create prompts for optimization, we constructed them around five representative loop optimization techniques: loop unrolling, loop jamming (also known as loop fusion), loop tiling, loop distribution (also known as loop fission), and loop interchange. The prompts comprise instructions and, optionally, include examples of specific optimization techniques. The type of prompts we investigated includes a zero-shot prompt, which comprises no examples, five one-shot prompts with each including one of four examples, and ten two-shot prompts, each comprising a permutation of example pairs together. In total, this resulted in 16 distinct prompts. In our study, we found that these types of prompts were effective, and most of the prompts were suitable for some of the benchmarks; however, it was not the case that one prompt was generally the best across the board. This generated a need to develop a recommender system that selects prompts and optimization examples based on the input code provided for optimization.

\subsubsection{Optimization Strategy Recommender}

The Recommender module functions as an integral sub-component of the overall optimization framework, tasked with identifying the most suitable optimization strategies and illustrative examples to include in the code-optimization prompt. To perform this selection, the Recommender employs a low-rank adaptation supervised-finetuned (LoRA-SFT) Qwen2.5-Coder-3B-Instruct large language model (LLM) \cite{team2024qwen2}. Given an input code fragment, the model is prompted to determine which, if any, of the five predefined loop-optimization techniques are applicable. In cases where the submitted code is already optimized with respect to looping constructs, the Recommender explicitly returns the label “No Loop Optimization Needed.”

Each of the five optimization strategies is paired with a corresponding example. Once the Recommender identifies the relevant strategies, the associated examples are assembled to form the few-shot context used by the downstream optimizer. Conversely, if the Recommender concludes that no further optimization is warranted, the system constructs a prompt instructing the optimizer to reproduce the original code without modification. This mechanism ensures that the optimization pipeline remains both targeted and conservative - providing guidance only when beneficial and preserving correctness when no additional improvement is advisable.

To fine-tune the Qwen2.5-Coder-3B-Instruct model for the task of recommending loop optimization strategies, we employed the LORE dataset \cite{lore_paper}, explicitly excluding the PolyBench/C subset to prevent overlap with downstream evaluation. The LORE corpus comprises approximately 70,000 code samples, each accompanied by program mutations and associated speedup metadata. From this collection, we retained only those mutated variants that achieved a performance improvement greater than 5\%, yielding a subset of approximately 17,000 candidate examples. To avoid bias toward any particular optimization technique, we drew a class-balanced sample from this filtered pool, ensuring an equal representation of the five loop optimization strategies.

Additionally, we extracted an equal number of samples from the unfiltered dataset where mutations resulted in performance regressions exceeding a 20\% slowdown. These were used to construct negative instances, examples in which no loop optimization should be applied, thereby enabling the model to distinguish between beneficial and detrimental transformation contexts. The resulting dataset contained 3,418 total examples, partitioned into 3,009 training samples and 409 validation samples.

Fine-tuning was performed using low-rank adaptation (LoRA), configured with a rank of 32 and a scaling factor (alpha) of 32, together with a LoRA-plus learning-rate ratio of 16. The model was trained for three epochs. This configuration was chosen to balance training stability, parameter efficiency, and the ability to capture the nuanced decision boundaries required for reliable optimization strategy recommendations.

\subsubsection{Feedback \& Prompt Synthesizer}

After the first transformation attempt, the feedback received from the verification chain (detailed below) is used to refine the optimizer's strategy. For example, if the verification feedback indicates that certain optimizations did not preserve correctness, the optimizer can modify its approach in subsequent attempts, enhancing the transformation process iteratively.


\subsection{Transform Verification}

The code transformations generated by the LLM-based Optimizer have the potential to be incorrect in several ways. Most importantly, the optimization attempt may have resulted in code that is functionally not equivalent to the original code. The goal of the verification chain is to assess the correctness of the transformed code using different approaches. Code transformations flagged as incorrect by the verification chain are rejected with an explanation (error messages, etc.) and sent back to the optimizer for another attempt. As demonstrated in the results, the verification chain increases the correctness of the overall optimization system.

The verification chain comprises three verification steps, each targeting distinct aspects of the code's correctness and functionality. These are: syntax verification, symbolic verification using the CBMC tool (bounded model checking) \cite{kroening2014cbmc}, and LLM-based verification. Note that we avoid test-based verification in the system, as it introduces runtime complexities that do not fit a compiler workflow (e.g., the need for setting up test cases, impact on workflow speed, etc.).


\subsubsection{Syntax Verifier}

The first step in the verification chain checks whether the optimized code adheres to correct syntax. This is performed by invoking the compiler with the \code{-fsyntax-only} flag. Syntax checking at the beginning of the chain helps root out clearly unusable transforms and fail early without the need for attempting more computationally expensive checks. In the case of syntax errors, the errors are communicated back to the optimizer for an attempt to fix the discovered problems.


\subsubsection{Symbolic Verification}

We explored two candidate tools for symbolic verification of transforms generated by the optimizer: Alive2 \cite{lopes2021alive2} and the C Bounded Model Checker (CBMC) \cite{kroening2014cbmc}.

\textbf{Alive2 \cite{lopes2021alive2}}. Alive2 provides a formal mechanism for verifying compiler optimizations through refinement checking, a  technique that validates the optimized program preserves the semantics of the original. Its approach encodes the original and optimized LLVM IR functions into SMT constraints and proves refinement by discharging them with efficient SMT solvers. Inspired by the use of Alive2 in the LLM-based code vectorization system described in \cite{taneja2025llm}, we experimented with Alive2 within our compiler workflow to validate the equivalence of LLM-generated optimizations at the LLVM-IR level.

\textbf{The C Bounded Model Checker (CBMC) \cite{kroening2014cbmc}}. CBMC is a widely used software verification tool that checks the properties of programs via symbolic execution within a bounded space. It translates a given C program with assertions (properties to be checked) into an intermediate representation with loops unrolled to a specified depth (bounded), and derives an SMT (Satisfiability Modulo Theories). The SMT formula is then checked for satisfiability with an efficient SMT solver, where satisfiability indicates violation of a property or assertion in the input program.

Prior works have demonstrated the potential of using CBMC for equivalence checking of programs. For example, it has been used for regression verification (proving that a modification to a program does not violate equivalence, which is highly useful for regression testing in software development) \cite{godlin2013regression}. In more recent work, CBMC has been used to detect faults (assertion failures) in Python code by first translating Python to C code using a large language model \cite{orvalho2025pyveritas}. However, to the best of our knowledge, our T-LLM Compiler is the first use case of CBMC for equivalence checking of C-level optimizations, particularly for LLM-generated optimizations.

To perform equivalence checking between the original function and the transformed or optimized function, a program needs to have the structure illustrated in Fig. \ref{fig:cbmc_code_outline}. The CBMC tool is invoked on this program as:

\noindent \code{cbmc --cvc5 --no-standard-checks --function main --unwind 20}

In this invocation, standard CBMC property checks are disabled (e.g. pointer and array bound checks etc.) as we are primarily interested in the equivalence assertion in the main function (\code{--function main}). The \code{--unwind} flag specifies the loop unrolling depth for performing bounded checking, which has to be reasonably small for tractable equivalence checking. The \code{cvc5} SMT solver is used in this work, as empirical results have shown that it is able to perform equivalence checking for larger loop unrolling depths than the default SMT solver in CBMC or the \code{Z3} SMT solver.

\begin{figure}[!ht]
\centering
    \begin{minipage}{0.95\linewidth}
        \begin{lstlisting}[language=C]
#include <assert.h>

#define N 15

int nondet_int();   // CBMC function

// Original function
void kernel_original(int n, int input[N], int output[N]) {
    for (int i = 0; i < n; i++)
        output[i] = input[i] * 2;
}

// Transformed function
void kernel_transformed(int n, int input[N], int output[N]) {
    for (int i = 0; i < n; i++)
        output[i] = input[i] + input[i];
}

int main() {
    int input1[N], output1[N];
    int input2[N], output2[N];
    
    // Initialize input arrays with non-deterministic values
    for (int i = 0; i < N; i++) {
        input1[i] = nondet_int();
        input2[i] = input1[i]; // Make identical copies
    }
    
    // Run both functions
    kernel_original(N, input1, output1);
    kernel_transformed(N, input2, output2);
    
    // Assert equivalence of outputs per array element
    for (int i = 0; i < N; i++)
        assert(output1[i] == output2[i]);
    
    return 0;
}
        \end{lstlisting}
    \end{minipage}
    \caption{C program with code augmentations necessary for function equivalence checking by CBMC}
    \label{fig:cbmc_code_outline}
\end{figure}


\subsection*{Limitations of CBMC for Equivalence Checking}

While CBMC is a very useful tool, it does come with some limitations. These include limited loop bounds, handling floating-point data types, and potential false equivalence outcomes. In particular:

\textbf{Limited loop bounds}. A loop unrolling depth or bound needs to be specified for each loop in the code, which means the CBMC equivalence assertion is validated for a constrained set of paths in the program (bounded model checking). This typically corresponds to limited input and output array sizes, as loops often iterate on arrays. When large array sizes are used, CBMC is unable to complete the validation in a reasonable amount of time. For example, on PolyBench/C, matrix sizes needed to be under 15 x 15 for successful CBMC verification.
    
\textbf{High computational complexity with floating point data types}. For code with complex operations on arrays (e.g. matrix operations such as those in PolyBench/C), the CBMC validation takes too long or fails when the array data type is float. This is likely due to the search space explosion caused by float data types and the complexities of the floating-point decision procedure in CBMC. On PolyBench/C, this forced us to use integer array elements, resulting in tractable and useful equivalence checking.

\textbf{Potential for false equivalence outcomes}. Equivalence verification on a limited input space means that a violation existing outside this space will not be detected by CBMC. In other words, a CBMC verification pass (decision: equivalent) does not guarantee that the two functions are equivalent. However, it remains a strong indication of equivalence, as corroborated by our empirical experiments.


\subsubsection{LLM-based Semantic Verifier}

While CBMC has been shown in our experiments to provide an improvement in the code optimization accuracy, it was unable to detect as many errors as we hoped. As such we had to supplement this symbolic verification with our own LLM-based verification strategy. In our approach, functional equivalence between the original and the transformed code is determined by prompting an LLM in the form shown in Fig. \ref{fig:llm_verifier_prompt_structure}. In this work, we use detailed prompts on pretrained and instruction-tuned LLMs without doing any finetuning, to evaluate the potential of base LLMs for the task for assessing equivalence between two versions of code. 

The prompting strategies explored in our work include:
\begin{itemize}
    \item Single independent prompt (Fig. \ref{fig:llm_verifier_prompt_single_detailed}). A generic prompt asking to check equivalence between the two versions of C code (original and transformed). Detailed instructions and hints are given on the important aspects of the code comparison (e.g. code paths, data dependencies).
    
    \item Three prompts dependent on the CBMC Verifier's decision. The CBMC Verifier can produce one of three decisions: accept, reject, non-decision (failures such as timeouts). Also, a rejection by CBMC is more trustworthy than an accept (due to constrained input search space). To reflect this, we prepare two additional prompts for the reject and accept decisions of the CBMC Verifier, by adding information about the CBMC decision to the prompt (Figs. \ref{fig:llm_verifier_prompt_cbmc_decision_accepted}, \ref{fig:llm_verifier_prompt_cbmc_decision_rejected}). In the case of a non-decision from the CBMC Verifier, the system resorts to the previous single independent prompt.
\end{itemize}

\begin{figure}[!ht]
\centering
    \begin{minipage}{0.95\linewidth}
        \begin{lstlisting}[]
Given below are two versions of a C function ... [more context] ...
Task: Evaluate whether the optimized version is functionally equivalent to the original.

Original function:
{ORIGINAL_CODE}

Optimized version:
{TRANSFORMED_CODE}

Instructions on elements to look for (e.g. code paths, data dependencies): {}
Output format instructions: {}
        \end{lstlisting}
    \end{minipage}
    \caption{Structure of the prompts used in the LLM Verifier}
    \label{fig:llm_verifier_prompt_structure}
\end{figure}


\section{Experiments and Results}

To test the effectiveness of the T-LLM Compiler we tested it on a set of 30 PolyBench/C benchmarks suite and compared it to the -O3 BiSheng Enterprise compiler. The runtime of each benchmark was measured on a 128C AArch64 ARM64 server with 512GB RAM. The two primary performance metrics used to evaluate the system are:

\begin{itemize}
    \item \textbf{Correctness}: The percentage of correctly transformed functions, as validated by a test suite.
    \item \textbf{Average speedup}: The average runtime performance improvement achieved by the optimizations across the dataset. Note that if an optimization produced by the system is incorrect, we treat its speedup as 1.0.
    \item \textbf{Average speedup of Correctly Transformed Kernels}: The average runtime performance improvement of kernels that were transformed and correct.
\end{itemize}

In addition to end-to-end testing, we examine the utility and effectiveness of each system component in more detail. This includes how we arrived at different prompting strategies and how they perform individually, challenges we faced with symbolic checker tools, and finally, the performance of different verification strategies.

We also executed our baseline using both the Bisheng and GCC compilers under the O2 and O3 optimization levels. As reported in Table \ref{tab:kernels_compilers}, the resulting runtimes exhibit no statistically significant differences, indicating comparable performance across compilers and optimization settings.

\begin{table*}[!htbp]
\centering
\caption{Kernels Execution Times in Seconds with Different Compilers and Flags}
\label{tab:kernels_compilers}
\resizebox{\textwidth}{!}{%
\begin{tabular}{lcccc}
\toprule
\textbf{Kernel} & \textbf{Bisheng with O3 Flag} & \textbf{Bisheng with O2 Flag} & \textbf{GCC with O3 Flag} & \textbf{GCC with O2 Flag} \\
\midrule
durbin          & 0.005                         & 0.005                         & 0.005                     & 0.005                     \\
lu              & 5.257                         & 5.258                         & 5.269                     & 5.283                     \\
gramschmidt     & 3.911                         & 3.975                         & 3.884                     & 3.941                     \\
ludcmp          & 5.257                         & 5.23                          & 5.243                     & 5.229                     \\
trisolv         & 0.005                         & 0.004                         & 0.005                     & 0.005                     \\
cholesky        & 2.711                         & 2.71                          & 2.712                     & 2.712                     \\
mvt             & 0.016                         & 0.016                         & 0.017                     & 0.017                     \\
doitgen         & 1.005                         & 1.005                         & 1.006                     & 1.007                     \\
3mm             & 5.503                         & 5.515                         & 5.497                     & 5.5                       \\
2mm             & 3.342                         & 3.334                         & 3.329                     & 3.33                      \\
bicg            & 0.027                         & 0.027                         & 0.027                     & 0.027                     \\
atax            & 0.009                         & 0.009                         & 0.012                     & 0.012                     \\
gesummv         & 0.012                         & 0.011                         & 0.012                     & 0.012                     \\
symm            & 2.502                         & 2.507                         & 2.532                     & 2.553                     \\
syr2k           & 2.899                         & 2.966                         & 2.137                     & 2.148                     \\
gemver          & 0.019                         & 0.019                         & 0.021                     & 0.021                     \\
syrk            & 1.302                         & 1.318                         & 0.892                     & 0.869                     \\
gemm            & 0.766                         & 0.756                         & 0.815                     & 0.769                     \\
trmm            & 1.419                         & 1.412                         & 1.419                     & 1.444                     \\
adi             & 17.659                        & 17.62                         & 17.107                    & 17.127                    \\
heat-3d         & 6.883                         & 6.848                         & 7.161                     & 7.15                      \\
jacobi-2d       & 3.595                         & 3.567                         & 4.173                     & 4.179                     \\
seidel-2d       & 32.433                        & 32.421                        & 32.356                    & 32.349                    \\
fdtd-2d         & 2.49                          & 2.45                          & 2.613                     & 2.624                     \\
jacobi-1d       & 0.002                         & 0.002                         & 0.002                     & 0.002                     \\
floyd-warshall  & 27.663                        & 29.622                        & 23.662                    & 23.626                    \\
deriche         & 0.212                         & 0.212                         & 0.215                     & 0.216                     \\
nussinov        & 4.057                         & 4.075                         & 4.257                     & 4.266                     \\
correlation     & 2.716                         & 2.663                         & 2.553                     & 2.556                     \\
covariance      & 2.697                         & 2.611                         & 2.61                      & 2.581                    \\ 
\bottomrule
\end{tabular}}
\end{table*}





\subsection{Optimizer Performance}

Table \ref{tab:opt_results} presents a comparative analysis of prompting strategies, examining their impact on kernel transformation success, functional correctness, and execution speedup. The baseline prompt, which contained no examples, achieved a 90\% success rate in kernel transformation and 80\% accuracy, accompanied by a modest average speedup of 1.167×. Incorporating one-shot prompts produced variable effects depending on the chosen example. For instance, One-shot Prompt 1 increased accuracy to 90\% but reduced transformation success to 86.67\%, while One-shot Prompt 2 maintained 90\% accuracy but yielded the lowest transformation rate at 73.33\%. One-shot Prompt 3 yielded a more balanced outcome, with 83.33\% transformation success and 86.67\% accuracy. Two-shot prompts provided greater stability across the evaluated metrics. Notably, combining Examples 2 and 3 resulted in the highest accuracy (96.67\%), albeit at a moderate transformation success rate of 80\%. In contrast, the pairing of Examples 3 and 4 yielded a more balanced performance, achieving an 86.67\% transformation, 90\% accuracy, and a reasonable speedup of 1.084×. These results suggest that carefully curated example combinations can enhance correctness while maintaining acceptable transformation reliability. Comparing one-shot prompts to two-shot prompts, we find that they have similar average accuracy; however, one-shot prompts show, on average, a higher overall speedup compared to two-shot prompts (11.83\% vs 9.37\%). One- and two-shot prompts are more accurate than the base prompt by approximately 7.5\%. This suggests that leveraging prompts with examples is more effective overall in producing a correct result; however, focusing on fewer classes of examples also provides more robust guidance to the LLM, enabling high-quality optimization to take effect.

When selecting the best-performing transformation for each kernel across all prompting strategies, the T-LLM compiler framework demonstrated 90\% transformation success and perfect functional correctness (100\%), alongside the highest observed speedup of 1.178×. This highlights the importance of prompt optimization in achieving both reliable correctness and improved efficiency.

Building upon these results, we constructed an optimization-strategy recommender system capable of autonomously selecting the most appropriate prompting configuration for each kernel. We evaluated the recommender under two conditions: with chain-of-thought (CoT) reasoning enabled and with CoT disabled. When CoT was employed, the system achieved a transformation success rate of 80\% and yielded a substantial average speedup of 1.115×. However, this improvement came with a reduction in accuracy, which was also measured at 80\%. Conversely, when CoT was omitted, the recommender achieved a slightly higher transformation success rate of 83.33\% and an average speedup of 1.079×, along with a modest gain in accuracy of 0.33\%. Although this comparison reveals a trade-off between the magnitude of speedup and predictive reliability, the overall findings indicate that automated prompt selection, irrespective of CoT usage, provides meaningful performance benefits while maintaining a high degree of transformation consistency.

Overall, the results highlight that the design of prompts has a significant impact on model performance. While individual one-shot and two-shot prompts illustrate the sensitivity of outcomes to example quality, the best results are achieved by carefully selecting optimal prompts for maximum accuracy and speedup or by employing a recommender system to enhance speed and consistency.

In our experimental evaluation, we employed three large language models (LLMs): CodeLlama-7B-Instruct, CodeLlama-13B-Instruct, and Qwen2.5-32B-Instruct. Table \ref{tab:opt_results_llms} summarizes the “Best of All Prompts” performance for each model. As the results indicate, Qwen2.5-32B-Instruct demonstrates a substantial performance advantage, markedly outperforming both CodeLlama variants across all reported metrics.

\begin{table*}[!htbp]
\captionof{table}{PolyBench/C results with different optimizer prompts using Qwen2.5 - 32B - Instruct as the LLM}
\label{tab:opt_results}
\begin{center}
\resizebox{\textwidth}{!}{
\begin{tabular}{lccccc}
\toprule
\textbf{Experiment} & \textbf{Examples}        & \textbf{Transformed Kernels} & \textbf{Accuracy} & \textbf{Average speedup} & \textbf{\begin{tabular}[c]{@{}c@{}}Average speedup of \\ Correctly Transformed Kernels\end{tabular}} \\
\midrule
Base Prompt             & No examples                 & 90.00\%                      & 80.00\%           & 1.167                     & 1.240                                                                                                 \\
One-shot Prompt 1       & Unrolling                   & 86.67\%                      & 90.00\%           & 1.091                     & 1.122                                                                                                 \\
One-shot Prompt 2       & Jamming                     & 73.33\%                      & 90.00\%           & 1.109                     & 1.175                                                                                                 \\
One-shot Prompt 3       & Tiling                      & 83.33\%                      & 86.67\%           & 1.112                     & 1.163                                                                                                 \\
One-shot Prompt 4       & Interchange                 & 76.67\%                      & 83.33\%           & 1.161                     & 1.267                                                                                                 \\
One-shot Prompt 5       & Distribution                & 76.67\%                      & 90.00\%           & 1.035                     & 1.055                                                                                                 \\
Two-shot Prompt 1       & Unrolling \& Jamming        & 83.33\%                      & 83.33\%           & 1.098                     & 1.114                                                                                                 \\
Two-shot Prompt 2       & Unrolling \& Tiling         & 86.67\%                      & 83.33\%           & 1.104                     & 1.149                                                                                                 \\
Two-shot Prompt 3       & Unrolling \& Interchange    & 83.33\%                      & 83.33\%           & 1.124                     & 1.186                                                                                                 \\
Two-shot Prompt 4       & Jamming \& Tiling           & 80.00\%                      & 96.67\%           & 1.098                     & 1.127                                                                                                 \\
Two-shot Prompt 5       & Jamming \& Interchange      & 80.00\%                      & 90.00\%           & 1.054                     & 1.080                                                                                                 \\
Two-shot Prompt 6       & Tiling \& Interchange       & 86.67\%                      & 90.00\%           & 1.084                     & 1.110                                                                                                 \\
Two-shot Prompt 7       & Distribution \& Unrolling   & 83.33\%                      & 76.67\%           & 1.069                     & 1.116                                                                                                 \\
Two-shot Prompt 8       & Distribution \& Tiling      & 83.33\%                      & 76.67\%           & 1.056                     & 1.092                                                                                                 \\
Two-shot Prompt 9       & Distribution \& Interchange & 83.33\%                      & 86.67\%           & 1.165                     & 1.237                                                                                                 \\
Two-shot Prompt 10      & Distribution \& Jamming     & 86.67\%                      & 66.67\%           & 1.092                     & 1.174                                                                                                 \\
Recommender without CoT & -                           & 83.33\%                      & 83.33\%           & 1.079                     & 1.120                                                                                                 \\
Recommender with CoT    & -                           & 80.00\%                      & 80.00\%           & 1.115                     & 1.194                                                                                                 \\
Best of All Prompts     & -                           & 90.00\%                      & 100.00\%          & 1.175                     & 1.195                                                                                                 \\
\bottomrule
\end{tabular}}
\end{center}
\end{table*}

\begin{table*}[!htbp]
\captionof{table}{Best PolyBench/C results with different LLMs}
\label{tab:opt_results_llms}
\begin{center}
\resizebox{\textwidth}{!}{%
\begin{tabular}{lcccc}
\toprule
\textbf{LLM Model} & \textbf{Transformed Kernels} & \textbf{Accuracy} & \textbf{Average speedup} & \textbf{\begin{tabular}[c]{@{}c@{}}Average speedup of \\ Correctly Transformed Kernels\end{tabular}} \\
\midrule
CodeLlama - 7B     & 6.67\%                       & 100\%             & 0.99                      & 0.99                                                                                                  \\
CodeLLama - 13B    & 10\%                         & 100\%             & 0.97                      & 0.74                                                                                                  \\
Qwen2.5 - 32B      & 90\%                         & 100\%             & 1.178                     & 1.198    \\
\bottomrule
\end{tabular}}
\end{center}
\end{table*}



\subsection{T-LLM Compiler Workflow in Action}

To demonstrate the flow, we elaborate on one of the examples as it was processed by our T-LLM compiler. This example is the {\it doitgen} benchmark in the PolyBench/C suite shown in Fig. \ref{fig:tllm_compiler_loop_example_a}.

The core computation of this kernel is found in the {\it q} loop where an array {\it sum} is essentially computing a vector inner product. The performance problem here is that while array A is accessed efficiently, array C4 is not. The solution to the performance problem is to interchange loops {\it p} and {\it s}, which will cause array A to be accessing a loop invariant value in the inner most loop {\it p}, but now array C4 will be accessing data sequentially.

The prompt used in this example is a one-shot prompt with a loop interchange example, which initially generates code in Fig. \ref{fig:tllm_compiler_loop_example_b}. This code contains an easy-to-identify error, where the final results are prematurely stored in array A. This error is detected by the CBMC verifier, which causes the T-LLM Compiler flow to reject this solution and attempt a second iteration to fix the problem. Based on a feedback prompt, the second attempt is made in which a correct transformation is applied and verified by CMBC as shown
in Fig. \ref{fig:tllm_compiler_loop_example_c}. This enables the BiSheng Enterprise compiler to effectively handle this loop and generate efficient code that runs 4.33X faster than the original.

It is noteworthy that this solution was only found by exactly one prompt that exclusively suggested interchange as a proposed avenue of optimization. In addition, we examined the LLVM-IR of the two versions of the code and found that an improved memory layout was only part of the reason for the speedup. The other part was the application of vectorization by a factor of 2 to the inner most {\it p} loop. Although the prompt did not request vectorization, and none was applied by the LLM, the loop interchange and associated code changes enabled the BiSheng compiler to recognize that vectorization was profitable in this case, causing it to be applied. This is a very clear example of LLM collaborating with the traditional compiler flow to overcome performance bottlenecks.

A summary of all results for one-shot prompt 4, from which this example is taken, is provided in Table \ref{tab:kernels_transformation_details}.

\begin{figure}[!ht]
\centering
    \begin{minipage}{0.95\linewidth}
    \begin{lstlisting}[language=C]
void kernel_doitgen(int nr, int nq, int np, int 3D_ARRAY(A,NR,NQ,NP,nr,nq,np), int 2D_ARRAY(C4,NP,NP,np,np), int 2D_ARRAY(sum,NP,np))
{
  int r, q, p, s;
  for (r = 0; r < _PB_NR; r++)
    for (q = 0; q < _PB_NQ; q++)  {
      for (p = 0; p < _PB_NP; p++)  {
        sum[p] = SCALAR_VAL(0.0);
        for (s = 0; s < _PB_NP; s++)
            sum[p] += A[r][q][s] * C4[s][p];
      }
      for (p = 0; p < _PB_NP; p++)
        A[r][q][p] = sum[p];
    }
}
    \end{lstlisting}
    \end{minipage}
    \caption{Original PolyBench kernel: doitgen}
    \label{fig:tllm_compiler_loop_example_a}
\end{figure}

\begin{figure}[!ht]
\centering
    \begin{minipage}{0.95\linewidth}
    \begin{lstlisting}[language=C]
void kernel_doitgen(int nr, int nq, int np, int 3D_ARRAY(A,NR,NQ,NP,nr,nq,np), int 2D_ARRAY(C4,NP,NP,np,np), int 2D_ARRAY(sum,NP,np))
{
  int r, q, p, s;
  for (r = 0; r < _PB_NR; r++) {
    for (q = 0; q < _PB_NQ; q++) {
      // Initialize sum array
      for (p = 0; p < _PB_NP; p++) {
        sum[p] = SCALAR_VAL(0.0);
      }
      // Optimization attempt: compute the sum and update A in a single loop to reduce memory access latency
      for (s = 0; s < _PB_NP; s++) {
        for (p = 0; p < _PB_NP; p++) {
          sum[p] += A[r][q][s] * C4[s][p];
          A[r][q][p] = sum[p];  // INCORRECT: premature assignment
        }
      }
    }
  }
}
    \end{lstlisting}
    \end{minipage}
    \caption{Attempt \#1 by the Optimizer: incorrect, rejected by the CBMC Verifier}
    \label{fig:tllm_compiler_loop_example_b}
\end{figure}

\begin{figure}[!ht]
\centering
    \begin{minipage}{0.95\linewidth}
\begin{lstlisting}[language=C]
void kernel_doitgen(int nr, int nq, int np, int 3D_ARRAY(A,NR,NQ,NP,nr,nq,np), int 2D_ARRAY(C4,NP,NP,np,np), int 2D_ARRAY(sum,NP,np))
{
  int r, q, p, s;
  for (r = 0; r < _PB_NR; r++) {
    for (q = 0; q < _PB_NQ; q++) {
      // Initialize sum array
      for (p = 0; p < _PB_NP; p++) {
        sum[p] = SCALAR_VAL(0.0);
      }
      // Compute the sum and update A in a single loop to reduce memory access latency
      for (s = 0; s < _PB_NP; s++) {
        for (p = 0; p < _PB_NP; p++) {
          sum[p] += A[r][q][s] * C4[s][p];
        }
      }
      // Update A with the computed sums
      for (p = 0; p < _PB_NP; p++) {
        A[r][q][p] = sum[p];
      }
    }
  }
}
    \end{lstlisting}
    \end{minipage}
    \caption{Attempt \#2 by the Optimizer: correct and fast (4.33x speedup)}
    \label{fig:tllm_compiler_loop_example_c}
\end{figure}

\begin{table*}[!htbp]
\centering
\caption{Kernels Transformation, Verification, Speedup, \& Correctness with One-shot Prompt 4}
\label{tab:kernels_transformation_details}
\resizebox{\textwidth}{!}{%
\begin{tabular}{lccccccc}
\toprule
\textbf{Kernel} & \textbf{Transformed} & \textbf{No. of Attempts} & \textbf{Syntax Verification} & \textbf{CBMC Verification} & \textbf{LLM Verification} & \textbf{Test-based Validation} & \textbf{speedup} \\ 
\midrule
durbin          & True                 & 1                       & True                         & True                       & True                      & True                           & 1.250             \\ 
lu              & False                & 1                       & True                         & True                       & True                      & True                           & 0.997             \\ 
gramschmidt     & False                & 1                       & True                         & True                       & True                      & True                           & 1.018             \\
ludcmp          & True                 & 1                       & True                         & True                       & True                      & True                           & 1.004             \\ 
trisolv         & False                & 1                       & True                         & True                       & True                      & True                           & 1.000             \\ 
cholesky        & False                & 1                       & True                         & True                       & True                      & True                           & 0.999             \\ 
mvt             & True                 & 1                       & True                         & True                       & True                      & True                           & 1.455             \\ 
doitgen         & True                 & 2                       & True                         & False                      & True                      & True                           & 4.332             \\ 
3mm             & False                & 1                       & True                         & True                       & True                      & True                           & 0.999             \\ 
2mm             & False                & 1                       & True                         & True                       & True                      & True                           & 1.003             \\ 
bicg            & True                 & 2                       & True                         & True                       & True                      & True                           & 1.000             \\ 
atax            & True                 & 1                       & True                         & True                       & True                      & True                           & 0.692             \\ 
gesummv         & True                 & 1                       & True                         & True                       & True                      & True                           & 1.500             \\ 
symm            & True                 & 1                       & True                         & True                       & True                      & True                           & 0.996             \\ 
syr2k           & True                 & 1                       & True                         & False                      & True                      & False                          & 1.000             \\ 
gemver          & True                 & 1                       & True                         & True                       & True                      & True                           & 1.188             \\ 
syrk            & True                 & 2                       & True                         & True                       & True                      & True                           & 1.303             \\ 
gemm            & True                 & 1                       & True                         & True                       & True                      & True                           & 1.028             \\ 
trmm            & True                 & 1                       & True                         & True                       & True                      & True                           & 1.054             \\ 
adi             & True                 & 1                       & True                         & False                      & True                      & False                          & 1.000             \\ 
heat-3d         & True                 & 1                       & True                         & False                      & True                      & False                          & 1.000             \\ 
jacobi-2d       & True                 & 1                       & True                         & False                      & True                      & False                          & 1.000             \\ 
seidel-2d       & True                 & 1                       & True                         & True                       & True                      & True                           & 1.000             \\ 
fdtd-2d         & True                 & 1                       & True                         & False                      & True                      & False                          & 1.000             \\ 
jacobi-1d       & True                 & 1                       & True                         & True                       & True                      & True                           & 0.667             \\ 
floyd-warshall  & True                 & 1                       & True                         & True                       & True                      & True                           & 1.218             \\ 
deriche         & True                 & 1                       & True                         & True                       & True                      & True                           & 0.986             \\ 
nussinov        & False                & 1                       & True                         & True                       & True                      & True                           & 0.994             \\ 
correlation     & True                 & 1                       & True                         & True                       & True                      & True                           & 1.137             \\ 
covariance      & True                 & 1                       & True                         & True                       & True                      & True                           & 1.001             \\ 
\bottomrule
\end{tabular}
}
\end{table*}

\subsection{Verification Chain Ablations}

To understand the contribution of individual verification components and their combinations, we conducted an ablation study. The goal of this study is to isolate the effect of different verification setups on system performance, both in terms of correctness and achieved speedup. The results are shown in Table \ref{tab:verifier_ablation_results}.

\textbf{Verification substantially improves correctness}. Adding more verification steps consistently increases the overall accuracy of optimized code produced by the system, as incorrect or non-equivalent transformations are rejected and retried by the optimizer. For example, accuracy improves from 60\% in the no-verification baseline to 87\% with the full verification chain.

\textbf{Importantly, the addition of verification does not hinder speedup}. If the verification is too strict or conservative, there is potential for good optimizations (correct and fast) to be rejected. The minimal false rejection rates of the verifiers in our system  allow valid optimizations to pass through, which preserves or slightly enhances speedup. A modest increase in average speedup is observed from 1.13 to 1.16. This indicates that catching and filtering incorrect optimizations, followed by re-attempts, yields not only correct but also faster optimized code.

While all three verification setups shown in the last three rows of Table \ref{tab:verifier_ablation_results} (combinations of CBMC Verifier and LLM Verifier) are suitable candidates for the T-LLM Compiler workflow, the configuration with Optimizer + Syntax Verifier + CBMC Verifier + LLM Verifier with 3 prompts (based on CBMC decision) consistently produces both high accuracy and speedup.

The ``Multiple attempts'' metric quantifies the proportion of kernels that required re-optimization due to verification rejections. Results show that only a small subset of kernels undergo re-attempts; however, these retries meaningfully contribute to the improved accuracy of the final outputs. This demonstrates that the verification chain effectively identifies incorrect optimizations without excessively penalizing the optimizer.

Finally, the ``Transformed kernels'' column reflects the fraction of kernels for which the system successfully produced a transformed version of the code. Since the system can, in principle, output the original code after exceeding the maximum number of attempts, this measure ensures that the pipeline is actually producing optimizations. Across all setups, the system achieves a consistently high proportion of transformations, with the maximum number of optimization attempts set to 3.

\begin{table*}[!htbp]
\centering
\small
\caption{System performance with different verification setups on PolyBench/C}
\label{tab:verifier_ablation_results}
\resizebox{\textwidth}{!}{
\begin{tabular}{lcccc}
\toprule
\textbf{Experiment}                                           & \textbf{Transformed kernels} & \textbf{Multiple attempts} & \textbf{Accuracy} & \textbf{Speedup} \\ 
\midrule
Optimizer only                                                & 100\%                        & 0\%                                                & 60\%              & 1.13             \\
Optimizer + Syntax Verifier                                   & 100\%                        & 0\%                                                & 73\%              & 1.14             \\
Optimizer + Syntax Verifier + CBMC                            & 90\%                         & 7\%                                                & 83\%              & 1.19             \\
Optimizer + Syntax Verifier + LLM (single independent prompt) & 97\%                         & 17\%                                               & 80\%              & 1.18             \\
Optimizer + Syntax Verifier + CBMC + 3-way LLM (3 prompts)    & 90\%                         & 10\%                                               & 87\%              & 1.16             \\
\bottomrule
\end{tabular}}
\end{table*}


\subsection{Classification Performance of the CBMC Verifier}

In the context of transform verification, type I errors correspond to false rejections of correct optimizations, thereby sacrificing opportunities for performance gains. Conversely, type II errors correspond to false acceptances of incorrect optimizations, which undermine correctness guarantees. The balance between these two error types is critical: reducing false rejections preserves potential speedups, while reducing false acceptances improves the correctness and trustworthiness of optimizations produced by the system.

Across 400 test cases (multiple runs on PolyBench/C with different prompting strategies), the CBMC Verifier reported an accuracy of 87\%, a false rejection rate of 8.4\% and a false acceptance rate of 33.8\%. The low false rejection rate makes a rejection by CBMC trustworthy (it indicates that a code path violating equivalence between the original and optimized function was found). However, the CBMC Verifier has a high false acceptance rate (does not detect many incorrect optimizations), which likely results from the constrained or bounded nature of the CBMC search.



\subsection{Practical Challenges in Alive2 Verification}

In our experiments, Alive2 did not prove effective for validating equivalence between original and optimized C-level functions. Even at modest loop unwind depths, the tool frequently encountered timeouts, failing to complete the refinement check. We attribute this to the complexity of PolyBench kernels, which often contain deep nested loops and intricate data dependencies. These structures translate into complex LLVM IR semantics and, consequently, highly non-trivial refinement relationships that Alive2 must reason about. Moreover, Alive2 was primarily designed as a translation validation framework for compiler-driven IR-level optimizations, with particular emphasis on correctly handling undefined behaviour in LLVM. In this context, its strengths do not directly align with verifying C-level transformations generated by our LLM-based optimizer, making it less suitable as a practical verification backend for our system.


\section{Conclusion}

In this paper, we presented a comprehensive framework for optimizing C programs, with a focus on loop-type programs from the PolyBench/C benchmark suite. We demonstrate that a combination of different types of tooling, including compilers, symbolic verifiers, and LLMs, can successfully coexist to provide high-quality optimization solutions. This is particularly evident in the 83\% success rate of code generation optimizations, which generally in prior art hovers around 50\% for code optimization tasks. While this success rate is high, the overall speedup of transformed programs reaches an impactful 26.7\% (one-shot prompt 4).

The main reason for this success is the effective utilization of a wide range of tools, including compilers, symbolic checkers and LLMs. This, in combination with the iterative flow that allows the framework to retry multiple times to successfully create a viable solution, is the key to success. In addition, one step that crucially differentiates us from competing approaches is that a traditional compiler is not removed from the flow of execution. In fact, the framework's output is presented to the BiSheng compiler as input for low-level optimization. The reason we did this stems from the realization that traditional compilers are highly effective in the tasks for which they were designed. What they cannot do is change the input source to unlock optimizations that a programmer would otherwise be able to apply. The T-LLM Compiler opens this path wide open with an accurate code generation approach and a collaboration with traditional compiler frameworks.

\section{Future Directions}

While the results presented in this paper are compelling, the natural question becomes: can we do better? The answer is a resounding yes. In our experiments, we found that there is a wider variety of prompting and optimization strategies we could apply to this set of benchmarks, and we have seen up to 52\% speedup possible if further advancements are made. To materialize these benefits, we need to solve a number of problems.

The first challenge, as previously discussed, lies in determining the most appropriate prompt for a given code segment. Prompt effectiveness can vary substantially across kernels and computational settings, making prompt selection a nontrivial component of the optimization pipeline. This variability highlights the need for a systematic investigation into how prompt design interacts with kernel characteristics and influences overall performance. While we introduced an intelligent prompt-recommendation mechanism capable of automatically proposing or selecting suitable prompts, its current performance indicates room for further enhancement. Improving this component would not only increase the reliability of prompt selection but also contribute to greater efficiency and scalability in future large-scale optimization systems.


A second observation is that, in our prompt analysis, we found one-shot prompting to be slightly more effective in terms of producing better speedups. This was intended to unlock optimization opportunities that the BiSheng compiler could then exploit to produce fast code. However, two-shot prompts performed well in certain examples as well, which perhaps suggests that rather than combining the request for two optimizations to be considered, sequential requests might be more effective. While this may not always be feasible and sometimes more complex multi-step optimizations are required using multi-shot prompts, at least for PolyBench/C-type benchmarks, this approach may be more productive and worth exploring.

Finally, there is the problem of handling large code bases. In the benchmarks we explored in this work, most of
the functions being optimized fit roughly under 50 lines of code; however, for larger functions that are more complex 
and longer, optimization remains a challenge. In addition, further optimization efforts will need to attempt to focus 
on program-wide transformations, which will face this kind of problem much faster. Solving this challenge, however, is 
how we believe we can reach level 4 in our vision for the T-LLM Compiler. 

\bibliographystyle{ACM-Reference-Format}
\bibliography{references}

\newpage
\onecolumn
\appendix

\section{Verification \& Optimization Prompts}

This section presents the prompt collection we used in our work.

\begin{figure}[!ht]
\centering
    \begin{minipage}{0.95\linewidth}
        \begin{lstlisting}[]
Given below are two versions of a C function that should compute the same result. Note that the second version was an attempt to optimize the original code, and the optimization may have broken the logic or the behavior of the original function (incorrect optimization). On the other hand, it may be a correctly optimized version as well. Your goal is to determine whether the optimized version of the code is functionally equivalent to the original version.

Original function:
{ORIGINAL_CODE}

Optimized version:
{TRANSFORMED_CODE}

Steps to verify functional equivalence:
1. Analyze the code paths/ flow and identify key operations and their order in both versions
2. Check for data dependencies and whether they are correctly preserved in the optimized version
3. Analyze boundary conditions and loop invariants
4. Compare mathematical equivalence of computations

Output format instructions:
- First, give your analysis of the the two code versions
- List of key similarities and differences detected
- Your verdict on functional equivalence: *MUST produce this output*: : [Only use the phrases: EQUIVALENT/ NOT_EQUIVALENT]
- Confidence level on your verdict: [HIGH/MEDIUM/LOW]
        \end{lstlisting}
    \end{minipage}
    \caption{Single independent LLM Verifier prompt}
    \label{fig:llm_verifier_prompt_single_detailed}
\end{figure}

\begin{figure}[!ht]
\centering
    \begin{minipage}{0.95\linewidth}
        \begin{lstlisting}[]
For further context, consider that the C Bounded Model Checker (CBMC) tool has asserted that the optimized code is equivalent to the original.
CBMC does this via symbolical execution by asserting that function calls to original and optimized versions produce the same output for a defined input space.
However, the input search space is small for computational feasibility (eg: small input arrays), so its 'equivalent' decision *cannot* be treated as final (it's possible that non-equivalence exists outside its search space).
  
Leverage the given context, but perform your own careful analysis
        \end{lstlisting}
    \end{minipage}
    \caption{LLM Verifier prompt with context on CBMC decision: accept (equivalent)}
    \label{fig:llm_verifier_prompt_cbmc_decision_accepted}
\end{figure}

\begin{figure}[!ht]
\centering
    \begin{minipage}{0.95\linewidth}
        \begin{lstlisting}[]
For further context, consider that the C Bounded Model Checker (CBMC) tool has asserted that the optimized code is not equivalent to the original.
CBMC does this via symbolical execution by asserting that function calls to original and optimized versions produce different outputs for some case of input within a defined input space.
This makes it highly likely that the optimized code is not equivalent to the original. However, the CBMC decision can also be incorrect for various practical reasons.

Leverage the given context, but perform your own careful analysis
        \end{lstlisting}
    \end{minipage}
    \caption{LLM Verifier prompt with context on CBMC decision: reject (not equivalent)}
    \label{fig:llm_verifier_prompt_cbmc_decision_rejected}
\end{figure}


\begin{figure}[!ht]
\centering
    \begin{minipage}{0.95\linewidth}
        \begin{lstlisting}[]
You are a helpful assistant specialized in optimizing C code.
Your goals are to improve the execution speed of the provided code by applying loop-level optimizations (loop unrolling, loop jamming, loop tiling, etc.) while preserving the logic of the code.
Rules to follow:
1. Preserve the logic of the code.
2. DO NOT change function signatures.
3. DO NOT define macros or include any header files.
4. DO NOT introduce calls to any undefined functions (e.g. min, max).
5. DO NOT use any external libraries.
6. DO NOT use any parallelization or concurrency libraries (e.g., OpenMP).
7. Do NOT use any compiler pragmas. In particular, any form of #pragma unroll(...), #pragma ivdep, #pragma omp, or any vendor-specific variant is strictly disallowed. If you include any #pragma, your answer is invalid.
8. DO NOT produce any line containing #pragma. This includes #pragma unroll(N) for any integer N. If a #pragma appears anywhere in your output, that output is invalid.
9. Return only valid C code with the requested transformations.
10. Enclose the final optimized code only between [OPT] and [/OPT] tags, with no extra text.

If you fail to follow any of the rules, the answer will be considered invalid.
        \end{lstlisting}
    \end{minipage}
    \caption{Optimizer system prompt}
    \label{fig:opt_system_prompt}
\end{figure}

\begin{figure}[!ht]
\centering
    \begin{minipage}{0.95\linewidth}
        \begin{lstlisting}[]
You are a specialized assistant for compilers and code optimization in C. You will help me optimize my C code. Follow these steps to optimize the given C code:
    1. Analyze the provided C code to understand its functionality.
    2. Optimize it to improve its execution running time by applying loop optimization techniques such as loop unrolling, loop jamming, loop fusion, loop tiling, or other relevant optimizations that can help reduce execution time. 
    3. Wrap your final optimized C code between [OPT] and [/OPT] tags. Only return the best optimized code.

Rules to follow:
    1. Put one and only one version of the optimized code between [OPT] and [/OPT] tags.
    2. Follow the steps that are defined in the prompt.
    3. Do NOT use any external libraries. 
    4. Do NOT change the provided function signature, name or arguments.
    5. Do NOT use undefined functions like min, max, or swap.
    6. Do NOT include any header files or define any macros.
    7. Do NOT write any function with the name "main".

Here is the original C code:
{ORIGINAL_CODE}
Return only the most optimized C code within [OPT] and [/OPT] tags.
        \end{lstlisting}
    \end{minipage}
    \caption{Optimizer zero-shot prompt}
    \label{fig:opt_base_prompt}
\end{figure}

\begin{figure}[!ht]
\centering
    \begin{minipage}{0.95\linewidth}
        \begin{lstlisting}[]
Below is a negative example, demonstrating that compiler pragmas (e.g., #pragma ivdep, #pragma unroll, etc.) are not allowed:
NEGATIVE EXAMPLE (Input):
[ORG]
void anotherNegativeExample(int *arr, int n) {
#pragma unroll(N)
    for (int i = 0; i < n; i++) {
        arr[i] += 2;
    }
}
[/ORG]

NEGATIVE EXAMPLE (Output):
[OPT]
// We remove any compiler directive, especially #pragma unroll(N).
// Instead, we manually unroll or keep the loop as is.
void anotherNegativeExample(int *arr, int n) {
    // Example: manually unroll by factor 2, if desired:
    int i;
    for (i = 0; i < n - 2; i += 2) {
        arr[i] += 2;
        arr[i+1] += 2;
    }
    for (; i < n; i++) {
        arr[i] += 2;
    }
}
[/OPT]
Below are several examples of unoptimized code (within [ORG] and [/ORG]) paired with their optimized versions (within [OPT] and [/OPT]).
{EXAMPLES}
Now, using the same approach, please optimize the following code.
Return only the optimized version between [OPT] and [/OPT] tags, with nothing else outside those tags:
[ORG]
{ORIGINAL_CODE}[/ORG]
        \end{lstlisting}
    \end{minipage}
    \caption{Optimizer few-shot prompts}
    \label{fig:opt_few_shot_prompt}
\end{figure}

\begin{figure}[!ht]
\centering
    \begin{minipage}{0.95\linewidth}
        \begin{lstlisting}[]
Below is your previously returned code, which failed verification, followed by the verifier's feedback. Please revise the code so that it addresses the verifier's concerns. Continue to follow the rules about preserving function signatures, avoiding macros or undefined functions, avoiding compiler pragmas, and preserving the original logic.
Here is the original code:
[ORG]
{ORIGINAL}
[/ORG]
Your previously returned code which failed verification:
[OPT]
{FAILED_CODE}
[/OPT]
Verifier's feedback:
{FEEDBACK}
Now, please produce a corrected version of the code that addresses the feedback, adheres to all the rules, and returns only the optimized C code between [OPT] and [/OPT] tags with no extra text.
        \end{lstlisting}
    \end{minipage}
    \caption{Optimizer feedback prompt}
    \label{fig:opt_feedback_prompt}
\end{figure}

\begin{figure*}[!htbp]
\centering
\begin{adjustbox}{height=0.17\textheight}
\begin{minipage}{0.5\textwidth}
\begin{lstlisting}[]
EXAMPLE 1 (Input):
[ORG]
void doubleElements(int *arr, int n) {
    for (int i = 0; i < n; i++) {
        arr[i] *= 2;
    }
}
[/ORG]
EXAMPLE 1 (Output):
Optimization Technique: Loop Unrolling
[OPT]
void doubleElements(int *arr, int n) {
    int i;
    for (i = 0; i <= n - 4; i += 4) {
        arr[i]   *= 2;
        arr[i+1] *= 2;
        arr[i+2] *= 2;
        arr[i+3] *= 2;
    }
    // Handle leftover elements
    for (; i < n; i++) {
        arr[i] *= 2;
    }
}
[/OPT]
\end{lstlisting}
\end{minipage}
\end{adjustbox}\hfill
\begin{adjustbox}{height=0.17\textheight}
\begin{minipage}{0.51\textwidth}
\begin{lstlisting}[]
EXAMPLE 2 (Input):
[ORG]
void processArrays(int *arr1, int *arr2, int n) {
    for (int i = 0; i < n; i++) {
        arr1[i] += 1;
    }
    for (int i = 0; i < n; i++) {
        arr2[i] += 2;
    }
}
[/ORG]
EXAMPLE 2 (Output):
Optimization Technique: Loop Jamming
[OPT]
void processArrays(int *arr1, int *arr2, int n) {
    for (int i = 0; i < n; i++) {
        arr1[i] += 1;
        arr2[i] += 2;
    }
}
[/OPT]
\end{lstlisting}
\end{minipage}
\end{adjustbox}

\vspace{0.3em}

\begin{adjustbox}{height=0.17\textheight}
\begin{minipage}{0.555\textwidth}
\begin{lstlisting}[]
EXAMPLE 3 (Input):
[ORG]
void addMatrices(int *A, int *B, int *C, int N) {
    for (int i = 0; i < N; i++) {
        for (int j = 0; j < N; j++) {
            C[i*N + j] = A[i*N + j] + B[i*N + j];
        }
    }
}
[/ORG]
EXAMPLE 3 (Output):
Optimization Technique: Loop Tiling
[OPT]
void addMatrices(int *A, int *B, int *C, int N) {
    int tileSize = 4;
    for (int i0 = 0; i0 < N; i0 += tileSize) {
        for (int j0 = 0; j0 < N; j0 += tileSize) {
            for (int i = i0; i < i0 + tileSize && i < N; i++) {
                for (int j = j0; j < j0 + tileSize && j < N; j++) {
                    C[i*N + j] = A[i*N + j] + B[i*N + j];
                }
            }
        }
    }
}
\end{lstlisting}
\end{minipage}
\end{adjustbox}\hfill

\caption{Optimizer examples used in few-shot prompts and recommender (1/2)}
\label{fig:optimizer_examples}
\end{figure*}

\begin{figure*}[!htbp]
\centering
\begin{adjustbox}{height=0.17\textheight}
\begin{minipage}{0.64\textwidth}
\begin{lstlisting}[]
EXAMPLE 4 (Input):
[ORG]
void sumCols(int *matrix, int *result, int rows, int cols) {
    for (int j = 0; j < cols; j++) {
        result[j] = 0;
    }
    for (int j = 0; j < cols; j++) {
        for (int i = 0; i < rows; i++) {
            result[j] += matrix[i*cols + j];
        }
    }
}
[/ORG]
EXAMPLE 4 (Output):
Optimization Technique: Loop Interchange
[OPT]
void sumCols(int *matrix, int *result, int rows, int cols) {
    // Initialize result
    for (int j = 0; j < cols; j++) {
        result[j] = 0;
    }
    // Interchange loop order for better memory locality 
    for (int i = 0; i < rows; i++) {
        for (int j = 0; j < cols; j++) {
            result[j] += matrix[i*cols + j];
        }
    }
}
[/OPT]
\end{lstlisting}
\end{minipage}
\end{adjustbox}

\vspace{0.3em}

\begin{adjustbox}{height=0.17\textheight}
\begin{minipage}{0.6\textwidth}
\begin{lstlisting}[]
EXAMPLE 5 (Input):
[ORG]
void processArrays(int a[], int b[], int c[], int size) {
    // One loop performs two independent operations
    for (int i = 0; i < size; i++) {
        b[i] = a[i] * 2;      // Task 1
        c[i] = a[i] + 10;     // Task 2
    }
}
[/ORG]
EXAMPLE 5 (Output):
Optimization Technique: Loop Distribution (Fusion)
[OPT]
void processArrays(int a[], int b[], int c[], int size) {
    // First loop: Task 1
    for (int i = 0; i < size; i++) {
        b[i] = a[i] * 2;
    }

    // Second loop: Task 2
    for (int i = 0; i < size; i++) {
        c[i] = a[i] + 10;
    }
}
[/OPT]
\end{lstlisting}
\end{minipage}
\end{adjustbox}
\caption{Optimizer examples used in few-shot prompts and recommender (2/2)}
\label{fig:optimizer_examples}
\end{figure*}

\begin{figure}[!ht]
\centering
    \begin{minipage}{0.95\linewidth}
        \begin{lstlisting}[]
You are a specialized assistant for compilers and code optimization in Given an Original Code snippet, without access to the Optimized Code, analyze the code and select optimization strategies that could be applied to improve performance from the list below:
- Loop Unrolling
- Loop Distribution(Fission)
- Loop Tiling
- Loop Interchange
- Loop Unroll and Jam
If the code is already optimal or no strategies are applicable, respond with "No Loop Optimization Needed".
Provide the list of applicable strategies in this format:
[STR]
- [Strategy 1]
- [Strategy 2]
...
- [Strategy N]
[/STR]
---
Now analyze the code:
{ORIGINAL_CODE}
#### Optimization Suggestions:
        \end{lstlisting}
    \end{minipage}
    \caption{Recommender strategy prompt}
    \label{fig:opt_recommender_user_prompt}
\end{figure}

\begin{figure}[!ht]
\centering
    \begin{minipage}{0.95\linewidth}
        \begin{lstlisting}[]
You are an expert code optimization assistant. Your task is to analyze the provided C code snippets and apply each given optimization strategy to enhance performance.
Here are the optimization strategies you can use:
{STRATEGIES}
Below are several examples of unoptimized code (within [ORG] and [/ORG]) paired with their optimized versions (within [OPT] and [/OPT]) that used the given strategies.
{EXAMPLES}
Now, using the same approaches to optimize the following code.
Return only the optimized version between [OPT] and [/OPT] tags, with nothing else outside those tags:
[ORG]
{ORIGINAL_CODE}
[/ORG]
        \end{lstlisting}
    \end{minipage}
    \caption{Optimizer prompt using recommender suggestions}
    \label{fig:opt_recommender_fs_prompt}
\end{figure}

\begin{figure}[!ht]
\centering
    \begin{minipage}{0.95\linewidth}
        \begin{lstlisting}[]
You are an expert code optimization assistant. Your task is to analyze the provided C code snippets and apply each given optimization strategy to enhance performance.
Here are the optimization strategies you can use:
{STRATEGIES}
Below are several examples of unoptimized code (within [ORG] and [/ORG]) paired with their optimized versions (within [OPT] and [/OPT]) that used the given strategies.
{EXAMPLES}
Explain how each optimization strategy has been applied in the examples above. Provide a brief description of the techniques used in each optimized code snippet.
        \end{lstlisting}
    \end{minipage}

\begin{minipage}{0.95\linewidth}
        \begin{lstlisting}[]
Use the following optimization strategies and the provided descriptions to optimize the given C code snippet. Wrap your optimized code between [OPT] and [/OPT] tags.
Here is the optimization strategies and their descriptions:
{STRATEGIES}
{EXPLANATION}
Here is the C code to optimize:
[ORG]
{ORIGINAL_CODE}
[/ORG]
Return only the optimized version between [OPT] and [/OPT] tags, with nothing else outside those tags.
        \end{lstlisting}
    \end{minipage}
    \caption{Optimizer prompt using recommender suggestions using chain of thought prompting}
    \label{fig:opt_recommender_fs_prompt}
\end{figure}

\end{document}